\documentclass[10pt,twocolumn,letterpaper]{article}

\usepackage{cvpr}
\usepackage{times}
\usepackage{epsfig}
\usepackage{graphicx}
\usepackage{amsmath}
\usepackage{amssymb}
\usepackage{bm}
\usepackage{bbm, dsfont}
\usepackage{microtype}
\usepackage{multirow}

\usepackage{algorithm}
\usepackage{algorithmic}
\usepackage{booktabs}
\usepackage{bm}
\usepackage{cases}

\usepackage[breaklinks=true,bookmarks=false]{hyperref}

\cvprfinalcopy 

\def\cvprPaperID{****} 

\begin{document}

\title{Multi-modal Feature Fusion with Feature Attention for VATEX Captioning Challenge 2020}

\author{Ke Lin\textsuperscript{1,2}, Zhuoxin Gan\textsuperscript{1}, Liwei Wang\textsuperscript{2}\\
	\textsuperscript{1}Samsung Research China - Beijing (SRC-B)\\ 
	\textsuperscript{2}School of EECS, Peking University, Beijing, China\\
	{\tt\small
		\{ke17.lin,zhuoxin1.gan\}@samsung.com, \{wanglw@pku.edu.cn\}}
}
\maketitle

\begin{abstract}
   This report describes our model for VATEX Captioning Challenge 2020. First, to gather information from multiple domains, we extract motion, appearance, semantic and audio features. Then we design a feature attention module to attend on different feature when decoding. We apply two types of decoders, top-down and X-LAN and ensemble these models to get the final result. The proposed method outperforms official baseline with a significant gap. We achieve 76.0 CIDEr and 50.0 CIDEr on English and Chinese private test set. We rank 2nd on both English and Chinese private test leaderboard.
\end{abstract}

\section{Introduction}

Video captioning is the task that generating a natural language description of a given video automatically and has drawn more and more attention on both academic and industrial communities. Encoder-decoder framework is the most commonly used structure in video captioning where a CNN is the encoder to gather multi-modal information and an RNN is the decoder to generate captions \cite{Venugopalan15}. Generally, a 2-D and/or a 3-D CNN network is used as encoder to extract the visual feature \cite{Aafaq19}. Regional feature \cite{Zhang19}, semantic feature \cite{Chen19} and audio feature \cite{Wang18} are also used in prior video captioning papers to boost the captioning performance. During training, ground truth is feed to the model, while when inference, predicted words are feed to the model, the previous word distribution is different for training and inference, and this is the so-called "exposure bias" problem \cite{Ranzato16}. To overcome this problem, scheduled sampling \cite{Bengio15} and reward optimization \cite{Rennie17} is the most frequently used approach. 

In this paper, to get information from multiple domains, we use two 3-D CNN networks to extract motion feature, a 2-D CNN network to extract appearance feature, a ECO network \cite{zolfaghari2018eco} to extract 2-D \& 3-D fusion feature, a semantic model to extract semantic feature and a audio network to extract audio feature. To fuse these features, we propose a feature attention module to give different features with different weights. We use two latest decoders, top-down model \cite{Anderson18} and X-LAN \cite{Yingwei20} model separately. We use a multi-stage training strategy to train the model with cross-entropy loss, word-level oracle and self-critical in turn. We ensemble the top-down model and X-LAN model to get the final captioning result.

\section{Methods}

\begin{figure*}[t]
	\begin{center}
		\includegraphics[width=1.0\linewidth]{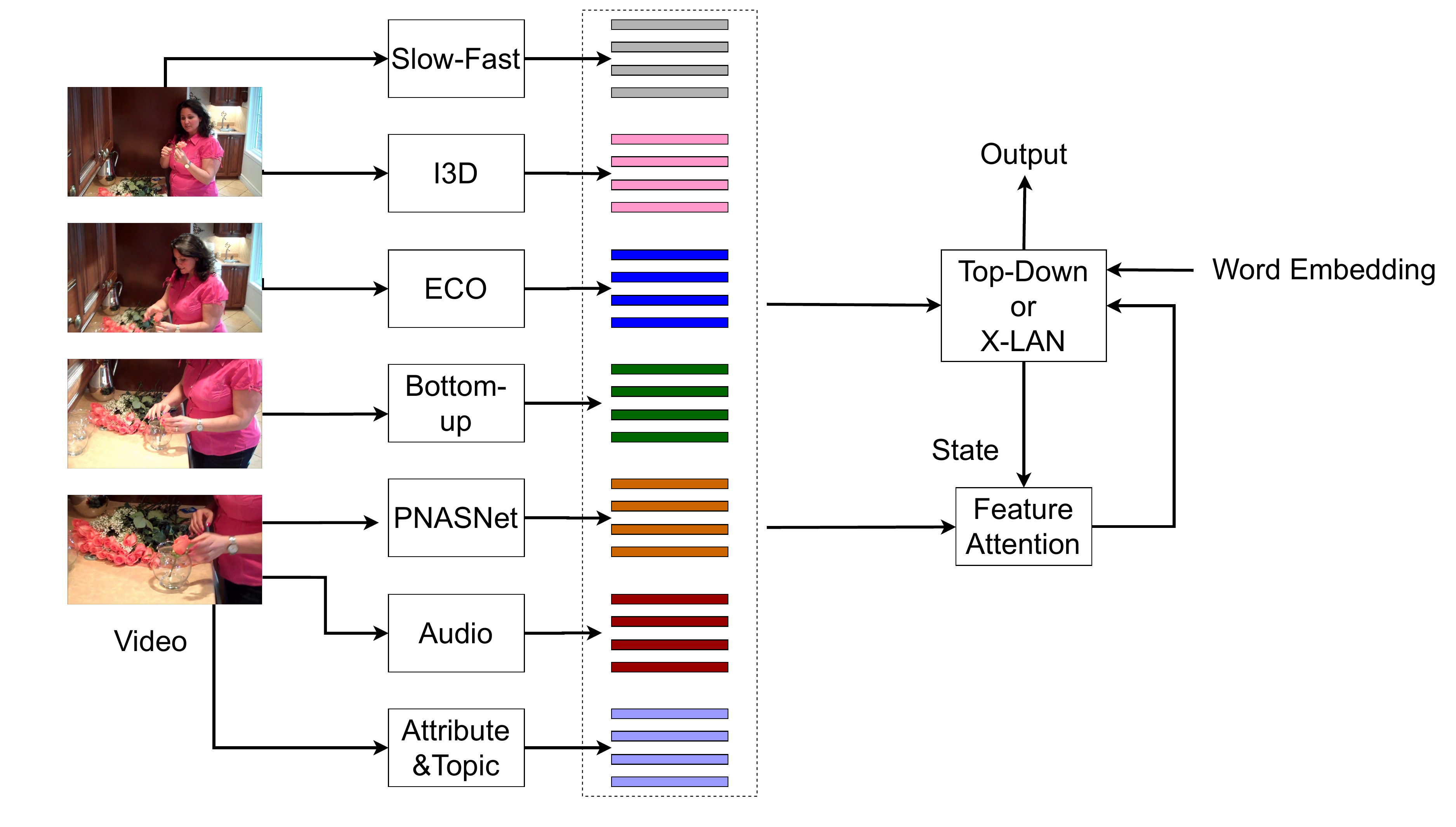}
	\end{center}
	\caption{The proposed video captioning framework.}
	\label{fig:long}
	\label{fig:onecol}
\end{figure*}

\subsection{Multi-modal Feature}

To enlarge the representation ability of the encoder, we extract multi-model feature from motion, appearance, semantic and audio domains. 

\noindent {\bf{Visual Feature.}}
In order to get motion feature of a given video, we use two types of 3-D CNN network. First, we use the I3D feature provided by the VATEX challenge organizers. Another 3-D CNN network is SlowFast \cite{Feichtenhofer19} network $\footnote{https://github.com/facebookresearch/SlowFast}$ pretrained on Kinetics-600 dataset, which has the identical data distribution of VATEX. In order to obtain the appearance feature, we use the PNASNet \cite{Liu18} $\footnote{https://github.com/Cadene/pretrained-models.pytorch}$ pretrained on ImageNet on every extracted frame. Inspired by bottom-up attention \cite{Anderson18}, local feature is also important for caption generation, so we also extract regional feature using a Faster R-CNN model pretrained on Visual Genome dataset for each extracted frame. Feature from different proposals are averaged to form the bottom-up feature for one frame. We also use an ECO \cite{zolfaghari2018eco} network pretrained on Kinetics-400 dataset to extract 2-D \& 3-D fusion feature. The above features are composed to the visual feature of a video. 

\noindent {\bf{Semantic Feature.}}
Inspired by some prior work \cite{Chen19, Aafaq19}, semantic prior is also helpful for video captioning. Following Chen's work \cite{Chen19}, we manually select the $300$ most frequent words from training and validation set as attributes. The semantic encoder consists of a multi-layer-perceptron on top of the visual featuers. We also extract topic ground truth from the training and validation set. Attribute and topic detection is treated as a multi-label classification task. Following \cite{Chen19}, we concatenate the predicted probability distribution of attributes/topic and the probability distribution of ECO, PNASNet and SlowFast as the semantic feature. Note that semantic feature is a one dimensional vector, we duplicate the semantic feature for $N$ times to align the semantic feature with the dimensions of other visual features, where $N$ is the number of frames.

\noindent {\bf{Audio Feature.}}
Inspired by Wang et al.'s work \cite{Wang18}, we also extract audio feature because audio from a video is also a powerful additional information. Our video feature extractor is a VGGish \cite{Hershey} network pretrained on Audioset dataset. First, we extract MEL-spectrogram patches for each audio. The sample rate of the audio is 16 KHz. The number of Mel filters is 64. The STFT window length is 25 ms and top length is 10 ms. 

All the above feature are embedded to the same dimension using fully connected layer.

\subsection{Decoder and feature attention}
We apply two types of decoders in this work, top-down model \cite{Anderson18} and X-LAN \cite{Yingwei20} model. Top-down consists of a two-layer GRU and an attention module. X-LAN employs a novel X-Linear attention block which fully leverages bilinear pooling to selectively focus on different visual information. We use a feature attention module to give feature with different weight. Denote the features with $\mathbf{V}_1,\dots,\mathbf{V}_n$, where $n$ is the number of the type of feature.

\begin{equation}
\mathbf{a}_t = \mathbf{softmax}(\mathbf{W}_{a}^{T}tanh(\mathbf{W}_{h}\mathbf{h}_{t-1}\bm{\mathbbm{1}}^T+\sum_{i=1}^n\mathbf{W}_{v,i}\mathbf{V}_i))
\end{equation}

\begin{equation}
\mathbf{V}_t = \sum^{n}_{i=1}a_{it}\mathbf{V}_i
\end{equation}

By the above equations, multi-modal feature are fused to a single feature $\mathbf{V}_t$. $\mathbf{h}_{t-1}$ is the hidden state of time $t-1$. For top-down model, $\mathbf{h}_{t-1}$ is the hidden state of the first GRU. For X-LAN model, $\mathbf{h}_{t-1}$ is the hidden state of the language LSTM.
\begin{table*}
	\centering
	\resizebox{\textwidth}{!}{
		\begin{tabular}{lrrrrrrrr}  
			\toprule
			\multirow{2}{*}{} & \multicolumn{4}{c}{\textbf{Validation}}  & \multicolumn{4}{c}{Test} \\
			\cmidrule(r){2-5} \cmidrule(r){6-9}
			\multirow{2}{*}{} & \textbf{BLEU-4} & \textbf{ROUGE-L} & \textbf{METEOR} & \textbf{CIDEr} & \textbf{BLEU-4} & \textbf{ROUGE-L} & \textbf{METEOR} & \textbf{CIDEr} \\
			\midrule
			Top-down & 0.377 & 0.525 & 0.255 & 0.882 & 0.337 & 0.502 & 0.237 & 0.716 \\
			X-LAN & 0.405 & 0.541 & \bf{0.267} & 0.885 & - & - & - & -  \\
			Ensemble & \bf{0.417} & \bf{0.543} & 0.265 & \bf{0.908} & \bf{0.392} & \bf{0.527} & \bf{0.250} & \bf{0.760} \\
			\bottomrule
	\end{tabular}}
	\caption{Comparison of captioning performance on VATEX English Captioning task.}
	\label{tab:booktabs}
\end{table*}

\begin{table*}
	\centering
	\resizebox{\textwidth}{!}{
		\begin{tabular}{lrrrrrrrr}  
			\toprule
			\multirow{2}{*}{} & \multicolumn{4}{c}{\textbf{Validation}}  & \multicolumn{4}{c}{Test} \\
			\cmidrule(r){2-5} \cmidrule(r){6-9}
			\multirow{2}{*}{} & \textbf{BLEU-4} & \textbf{ROUGE-L} & \textbf{METEOR} & \textbf{CIDEr} & \textbf{BLEU-4} & \textbf{ROUGE-L} & \textbf{METEOR} & \textbf{CIDEr} \\
			\midrule
			Top-down & 0.341 & 0.501 & 0.312 & 0.656 & \bf{0.331} & 0.495 & 0.301 & 0.479 \\
			X-LAN & 0.321 & 0.495 & 0.314 & 0.662 & - & - & - & -  \\
			Ensemble & \bf{0.341} & \bf{0.503} & \bf{0.315} & \bf{0.676} & 0.330 & \bf{0.497} & \bf{0.303} & \bf{0.504} \\
			\bottomrule
	\end{tabular}}
	\caption{Comparison of captioning performance on VATEX Chinese Captioning task.}
	\label{tab:booktabs}
\end{table*}

\subsection{Training strategy}

We train the video captioning model with three stages. In the first stage, we use the traditional cross-entropy loss for 5 epochs. The learning rate is fixed at $5\times10^{-4}$. Then we leverage word-level oracle \cite{Zhangwen19} with learning rate of $5\times10^{-5}$. When the CIDEr score of the validation is no longer growing for 2 epochs, we begin the third stage, self-critical policy gradient training. CIDEr and BLEU-4 are equally optimized. First, the learning rate is $5\times10^{-5}$. When the increase of CIDEr metric is saturated, we decreased the learning rate to $5\times10^{-6}$ to train the model until convergence.

\section{Results}

\subsection{Dataset and Preprocessing}
\noindent {\bf{VATEX.}} \cite{Wangxin19} VATEX contains over 41250 video clips in 10 seconds and each video clip depicts a single activity. Each video clip has 10 English descriptions and 10 Chinese descriptions. We use the official 25991 training examples as training data and 3000 validation examples for validation. 

We follow the standard caption pre-processing procedure including converting all words to lower cases, tokenizing on white space, clipping sentences over 30 words and filtering words which occur at least five times. We use open source Jieba $\footnote{https://github.com/fxsjy/jieba}$ toolbox to segment the Chinese words. The final vocabulary sizes are 10260 for VATEX English task, 12776 for VATEX Chinese task. We use standard automatic evaluation metrics including BLEU, METEOR, ROUGE-L and CIDEr.

We uniformly sample 32 frames for each video clip. The embedding dimension $512$. For top-down model, the model size and all hidden size are $512$. For X-LAN decoder, the model dimension is $512$. We train the captioning model using an Adam optimizer.

\subsection{Quantitative Result}

Table 1. and Table 2. show the quantitative result of the English and Chinese captioning tasks. For both English and Chinese captioning, X-LAN has better performance than top-down on validation set. The ensemble result achieves 0.76 CIDEr for English test set and 0.504 for Chinese test set. This result ranks 2nd on both English and Chinese captioning private test leaderboard.

\section{Conclusion}

In this challenge report, we propose a multi-modal feature fusion method. We gather feature from spatial, temporal, semantic and audio domains. We propose a feature attention module to attend on different feature when decoding. We use two latest captioning model: top-down and X-LAN. We use multiple stage training strategy to train the model. We rank 2nd at the official private test leaderboard for both English and Chinese captioning challenge.

{\small
\bibliographystyle{ieee_fullname}
\bibliography{egbib}

\begin{thebibliography}{10}\itemsep=-1pt

\bibitem{Aafaq19}
Nayyer Aafaq, Naveed Akhtar, Wei Liu, Syed~Zulqarnain Gilani, and Ajmal Mian.
\newblock Spatio-temporal dynamics and semantic attribute enriched visual
  encoding for video captioning.
\newblock In {\em {CVPR}}, 2019.

\bibitem{Anderson18}
Peter Anderson, Xiaodong He, Chris Buehler, Damien Teney, Mark Johnson, Stephen
  Gould, and Lei Zhang.
\newblock Bottom-up and top-down attention for image captioning and visual
  question answering.
\newblock In {\em CVPR}, 2018.

\bibitem{Chen19}
Haoran Chen, Ke Lin, Alexander Maye, Jianmin Li, and Xiaolin Hu.
\newblock A semantics-assisted video captioning model trained with scheduled
  sampling.
\newblock {\em arXiv preprint arXiv:1909.00121}, 2019.

\bibitem{Feichtenhofer19}
Christoph Feichtenhofer, Haoqi Fan, Jitendra Malik, and Kaiming He.
\newblock Slowfast networks for video recognition.
\newblock In {\em ICCV}, 2019.

\bibitem{Hershey}
Shawn Hershey, Sourish Chaudhuri, Daniel P.~W. Ellis, Jort~F. Gemmeke, Aren
  Jansen, R.~Channing Moore, Manoj Plakal, Devin Platt, Rif~A. Saurous, Bryan
  Seybold, Malcolm Slaney, Ron~J. Weiss, and Kevin~W. Wilson.
\newblock Cnn architectures for large-scale audio classification.
\newblock In {\em CoRR}, 2016.

\bibitem{Liu18}
Chenxi Liu1, Barret Zoph, Maxim Neumann, Jonathon Shlens, Wei Hua, Li-Jia Li,
  Li Fei-Fei, Alan Yuille, Jonathan Huang, and Kevin Murphy.
\newblock Progressive neural architecture search.
\newblock In {\em ECCV}, 2018.

\bibitem{Ranzato16}
Michael~Auli Marc’Aurelio~Ranzato, Sumit~Chopra and Wojciech Zaremba.
\newblock Sequence level training with recurrent neural networks.
\newblock In {\em {ICLR}}, 2016.

\bibitem{Yingwei20}
Yingwei Pan, Ting Yao, Yehao Li, and Tao Mei.
\newblock X-linear attention networks for image captioning.
\newblock In {\em CVPR}, 2020.

\bibitem{Rennie17}
Steven~J. Rennie, Etienne Marcheret, Youssef Mroueh, Jerret Ross, and Vaibhava
  Goel.
\newblock Self-critical sequence training for image captioning.
\newblock In {\em {CVPR}}, 2017.

\bibitem{Bengio15}
Navdeep~Jaitly Samy~Bengio, Oriol~Vinyals and Noam Shazeer.
\newblock Scheduled sampling for sequence prediction with recurrent neural
  networks.
\newblock In {\em {NeurIPS}}, 2015.

\bibitem{Venugopalan15}
Subhashina Venugopalan, Marcus Rohrbach, Jeff Donahue, Raymond Mooney, Trevor
  Darrell, and Kate Saenko.
\newblock Sequence to sequence - video to text.
\newblock In {\em {ICCV}}, 2015.

\bibitem{Wang18}
Xin Wang, Yuan-Fang Wang, and William~Yang Wang.
\newblock Watch, listen, and describe: Globally and locally aligned cross-modal
  attentions for video captioning.
\newblock {\em arXiv preprint arXiv:1804.05448}, 2018.

\bibitem{Wangxin19}
Xin Wang, Jiawei Wu, Junkun Chen, Lei Li, Yuan-Fang Wang, and William~Yang
  Wang.
\newblock Vatex: A large-scale, highquality multilingual dataset for
  video-and-language research.
\newblock In {\em ICCV}, 2019.

\bibitem{Zhang19}
Junchao Zhang and Yuxin Peng.
\newblock Object-aware aggregation with bidirectional temporal graph for video
  captioning.
\newblock In {\em {CVPR}}, 2019.

\bibitem{Zhangwen19}
Wen Zhang, Yang Feng, Fandong Meng, Di You, and Qun Liu.
\newblock Bridging the gap between training and inference for neural machine
  translation.
\newblock In {\em ACL}, 2019.

\bibitem{zolfaghari2018eco}
Mohammadreza Zolfaghari, Kamaljeet Singh, and Thomas Brox.
\newblock Eco: Efficient convolutional network for online video understanding,
  2018.
\newblock {Proceedings} of the European misc on Computer Vision (ECCV).

\end{thebibliography}
}

\end{document}